\documentclass{article}
\usepackage{spconf,amsmath,graphicx,float}

\usepackage{booktabs}
\usepackage{xcolor}
\usepackage{pifont}
\usepackage{wrapfig}
\usepackage{url}

\newcommand{\cmark}{\textcolor{cyan}{\ding{51}}}  
\newcommand{\xmark}{\textcolor{red}{\ding{55}}} 

\usepackage{listings}
\usepackage{xcolor}
\usepackage{caption}
\usepackage{cite}

\lstdefinelanguage{yaml}{
  keywords={concept_token, subject, style, settings},
  keywordstyle=\color{blue}\bfseries,
  basicstyle=\ttfamily\small,
  commentstyle=\color{gray},
  stringstyle=\color{teal},
  sensitive=false
}

\title{\textit{StoryState}: Agent-Based State Control for Consistent and Editable Storybooks}
\name{Ayushman Sarkar$^{1}$, Zhenyu Yu$^{2,*}$, Wei Tang$^{3}$, Chu Chen$^{3}$, Kangning Cui$^{4}$, Mohd Yamani Idna Idris$^{2}$}
\address{
${^1}$ Birbhum Institute of Engineering and Technology, 
${^2}$ Universiti Malaya, \\
${^3}$ City University of Hong Kong, 
${^4}$ Wake Forest University\\
}

\begin{document}
%
\maketitle
%
\begin{abstract}
Large multimodal models have enabled one-click storybook generation, where users provide a short description and receive a multi-page illustrated story. However, the underlying story state, such as characters, world settings, and page-level objects, remains implicit, making edits coarse-grained and often breaking visual consistency. We present \textit{StoryState}, an agent-based orchestration layer that introduces an explicit and editable story state on top of training-free text-to-image generation. \textit{StoryState} represents each story as a structured object composed of a character sheet, global settings, and per-page scene constraints, and employs a small set of LLM agents to maintain this state and derive 1Prompt1Story-style prompts for generation and editing. Operating purely through prompts, \textit{StoryState} is model-agnostic and compatible with diverse generation backends. System-level experiments on multi-page editing tasks show that \textit{StoryState} enables localized page edits, improves cross-page consistency, and reduces unintended changes, interaction turns, and editing time compared to 1Prompt1Story, while approaching the one-shot consistency of Gemini Storybook. Code is available at https://github.com/YuZhenyuLindy/StoryState
\end{abstract}
\section{Introduction}

Recent advances in large multimodal models have enabled ``one-click'' story creation in consumer applications. Systems such as Gemini Storybook can transform a brief textual description, optionally enriched with images, into a multi-page illustrated and narrated story in a single forward pass \cite{google2025storybook,google2025storybookblog,lin2026narratology,yu2025cotextor}, largely eliminating the need for manual storyboarding, prompt engineering, and layout design.  
Despite its convenience, one-click systems remains inherently \emph{one-shot} and \emph{version-based}. The underlying story structure, including character attributes, world settings, and per-page composition, remains implicit and inaccessible. Consequently, user edits are typically issued as natural-language requests that trigger a full regeneration, often causing unintended visual changes across multiple pages \cite{google2025storybookhelp}. Given the imperfect visual consistency of current image generators, even minor edits may cause cumulative cross-page drift, such as inconsistent character appearances or missing props \cite{theverge2025storybook,lin2026narratology}. We refer to this challenge as \emph{story-state control}: enabling \emph{precise, page-level edits} while preserving \emph{cross-page visual consistency}.

Prior efforts to improve consistency in visual storytelling can be broadly categorized into model-level and training-free approaches. \emph{Model-level personalization methods}, such as Textual Inversion, DreamBooth, SuTI, The Chosen One, and StoryMaker, achieve subject consistency by fine-tuning the generation model on specific characters or styles \cite{gal2022textual,ruiz2023dreambooth,chen2023suti,avrahami2024chosen,zhou2024storymaker}. \emph{Training-free methods}, including ConsiStory, StoryDiffusion, and One-Prompt-One-Story (1Prompt1Story), improve consistency through carefully engineered prompts and inference-time controls \cite{tewel2024consistory,zhou2024storydiffusion,liu2025onepromptonestory,li2025consistentstory,dong2025vista,wang2025storyanchors}. While effective for generating consistent visual narratives from well-crafted prompts, these methods operate primarily at the \emph{prompt level} and offer limited support for \emph{iterative, localized edits}, which are essential for refining multi-page stories \cite{he2025dreamstory,mao2024storyadapter,kwon2025dreamingcomics}.

In this work, we propose \textit{StoryState}, an agent-based orchestration layer that augments one-click story generation with an explicit, persistent story-state abstraction. \textit{StoryState} represents a story through three structured components: (i) a character sheet capturing persistent visual attributes, (ii) global world and style settings, and (iii) per-page scene constraints that define specific visual elements. A small set of LLM agents maintains and updates this state, derives structured prompts for generation and editing, and verifies consistency between outputs and the stored state. By externalizing story structure from the generative models, \textit{StoryState} enables precise page-level edits while preserving global visual consistency, without modifying or retraining any underlying models.
Our main \textbf{contributions} are:
\vspace{-6pt}
\begin{itemize}
    \item We introduce \textbf{an explicit story-state representation} that enables localized page-level edits while preserving cross-page visual consistency;
    \vspace{-7pt}
    \item We propose \textbf{an agent-based orchestration pipeline} that separates global identity from page-specific content for consistent generation and editing;
    \vspace{-7pt}
    \item We present \textbf{\textit{StoryState}} as a model-agnostic, training-free framework that interacts with multimodal backends solely through structured prompts.
\end{itemize}

\begin{figure*}[t]
  \centering
  \includegraphics[width=1.0\linewidth]{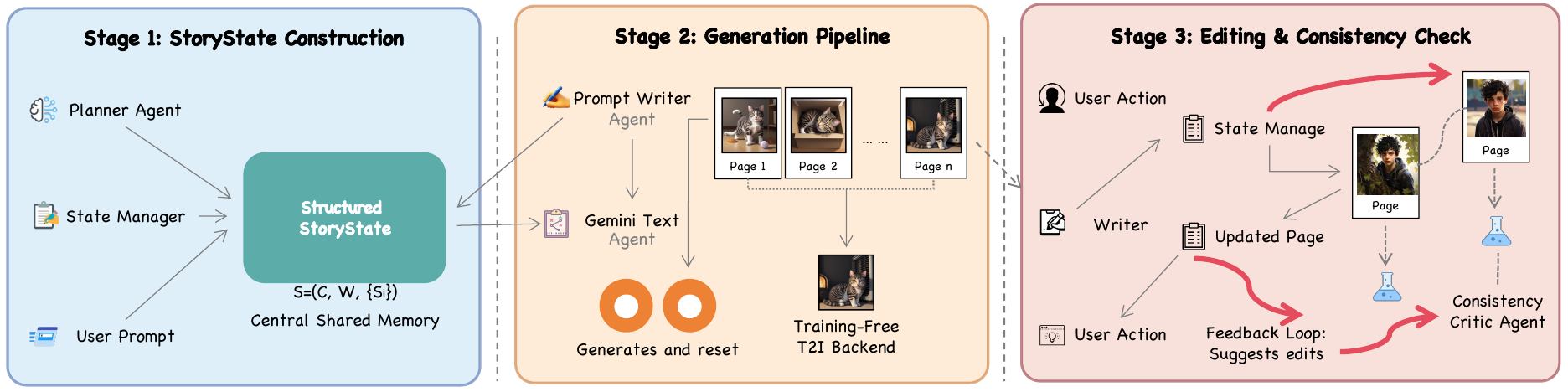}
  \caption{Overview of the \textit{StoryState} workflow. The system begins with a user prompt and constructs an explicit story state that comprises a character sheet, global world settings, and per-page scene descriptions. A team of LLM agents iteratively builds and updates this state, which drives both text and image generation through structured prompts. By maintaining this state persistently, \textit{StoryState} enables localized editing and cross-page consistency without modifying the underlying models.}
  \label{fig:overview}
\end{figure*}

\section{Related Work}
\noindent\textbf{Visual Storytelling Systems.}
Recent multimodal systems support one-click storybook generation from a single prompt \cite{google2025storybook,google2025storybookblog}. These systems provide end-to-end generation pipelines that abstract away authoring complexity. While convenient, these systems typically operate in a one-shot, version-based manner and offer limited control over intermediate representations such as characters, layout, or narrative flow \cite{google2025storybookhelp}. 

\noindent\textbf{Consistent Text-to-Image Generation.}
Maintaining visual consistency in multi-image generation has been studied through personalized methods (e.g., Textual Inversion, DreamBooth) that fine-tune models on character-specific inputs \cite{gal2022textual,ruiz2023dreambooth}, as well as training-free methods like ConsiStory and StoryDiffusion that manipulate prompts or generation constraints at inference time \cite{tewel2024consistory,zhou2024storydiffusion}. While effective at stabilizing visual attributes within static narratives, these techniques assume fixed prompts and offer little support for structured story updates.

\noindent\textbf{Interactive and Agent-Based Generation.}
Recent advances in interactive generation and plan-and-refine pipelines explore user control and coherence through iterative reasoning or agent-based decomposition \cite{gong2023interactive,lu2025multimodalplan}. However, existing approaches rarely maintain an explicit and persistent representation of the story's evolving state, making it hard to track consistency across edits or maintain coherence over time.

\section{\textit{StoryState}: Agent-Based State Control}

\textit{StoryState} introduces an explicit and editable story state maintained by a small set of LLM-based agents on top of storybook-style text generation and training-free text-to-image (T2I) backends. Instead of encoding story structure implicitly in model activations or prompts, \textit{StoryState} represents characters, world settings, and page-level scenes as a persistent state that serves as the single source of truth for text and image generation. Agents read from and write to this state to enable state-driven generation and localized editing. While our prototype uses a Gemini-based text agent and a 1Prompt1Story-style image pipeline, \textit{StoryState} is model-agnostic and compatible with any prompt-based text or image generation backend without retraining.

\subsection{\textit{StoryState} Construction}
\label{subsec:construction}

\noindent\textbf{Story-state representation.}
Each story is represented by a structured state
\begin{equation}
S = (C, W, \{S_i\}_{i=1}^N),
\end{equation}
where $N$ is the number of pages. The \emph{character sheet} $C$ is a collection of character entries, each storing (i) a name and narrative role, (ii) persistent visual attributes (e.g., species, age, appearance, clothing), and (iii) optional reference images. The world settings $W$ encode global elements shared across pages, including style, tone, and recurring locations or props. Each page state $S_i$ encodes page-specific information, consisting of (i) a brief scene description, (ii) the characters on that page (linked to $C$), (iii) explicit visual constraints (e.g., ``same yellow raincoat as on page~1'' or ``TV on the left''), and (iv) pointers to associated text and image assets.

Internally, $S$ is stored as a JSON-like structure that is inspectable and modifiable by agents. Crucially, this representation supports \emph{localized state updates}: edits are expressed as modifications to a minimal subset of $(C, W, \{S_i\})$, while all unrelated components remain unchanged. In contrast to the default Gemini Storybook workflow, where such information remains implicit within the model and is regenerated during each revision, \textit{StoryState} maintains this representation explicitly and persistently across editing cycles~\cite{google2025storybook,google2025storybookhelp}.

\noindent\textbf{Planner agent.}
Given an initial user prompt (e.g., ``a shy boy finds a lost robot in the city''), the Planner agent decomposes it into a page-level outline. This outline specifies a sequence of $N$ pages with scene descriptions, high-level narrative flow (e.g., beginning, climax, resolution), and initial character-page assignments. The Planner outputs a machine-readable structure that initializes the page states $\{S_i\}$, which provides a basis for downstream generation and editing.

\noindent\textbf{State Manager agent.}
The State Manager ingests the planner output and constructs consistent character and world representations, $C$ and $W$. When multiple references in the outline (e.g., ``a boy'', ``Tim'', and ``the child'') correspond to the same entity, the State Manager resolves them into a unified character entry and assigns stable visual attributes. It also records identity invariants, such as ``Tim always wears a yellow raincoat unless explicitly changed'', which are treated as persistent constraints during generation and editing.
This design decouples global identity information, stored once in $C$, from page-specific scene semantics, stored in $\{S_i\}$. By normalising ambiguous or redundant references into a canonical state representation, the State Manager ensures that identity attributes remain consistent across pages while allowing page-level semantics to vary independently. The unified story state $S$ becomes the central object that all agents read from and write to throughout generation and editing.

\subsection{Generation Pipeline}
\label{subsec:generation}

Once the story state $S$ is constructed, text and image generation proceed through two coordinated branches that share the same state representation. Both branches treat generation as a deterministic mapping conditioned on $S$ to ensure that identical states yield consistent outputs.

\noindent\textbf{Text agent.}
For each page $i$, the text agent consumes the page state $S_i$ along with the global context $(C, W)$ and generates the final narration for that page. During initial story creation, this process is applied to all pages $i = 1, \dots, N$. During editing, if a page state $S_i$ is modified (e.g., to adjust tone or length), the text agent regenerates only the affected page, while the narration of all other pages remains unchanged. This selective regeneration guarantees that textual edits remain localised and predictable.


\noindent\textbf{Prompt Writer agent and T2I backend.}
For image generation, \textit{StoryState} interfaces with a training-free T2I backend that supports 1Prompt1Story-style prompting~\cite{tewel2024consistory,zhou2024storydiffusion,liu2025onepromptonestory}. The Prompt Writer agent maps the full state $S = (C, W, \{S_i\})$ into a set of structured prompts comprising a global identity prompt $P_0$ and page-specific prompts $\{P_i\}_{i=1}^N$. The backend applies techniques such as Singular-Value Reweighting and Identity-Preserving Cross-Attention to strengthen identity features while preserving per-frame semantics~\cite{liu2025onepromptonestory}. By centralising shared identity information in $P_0$ and isolating scene-specific content each $P_i$, \textit{StoryState} reduces semantic redundancy across pages and mitigates global visual drift.

\subsection{Editing and Consistency Check}
\label{subsec:editing}

\noindent\textbf{Local and global editing via state updates.}
Unlike systems that rely on full-story regeneration for any change~\cite{google2025storybookhelp}, \textit{StoryState} performs edits by modifying only the relevant components of the state $S$. For a localized visual revision on page $j$ (e.g., ``on page~3, Lily should wear the same yellow coat as on page~1, and the TV should be on the left''), the State Manager updates $S_j$ to reflect the new constraints. The Prompt Writer then recomputes only the corresponding page prompt $P_j$, while keeping $P_0$ and all other $P_i$ unchanged, and the T2I backend regenerates the image for page $j$.

For global identity edits (e.g., ``Lily has green eyes throughout the story''), the State Manager updates the character sheet $C$. The Prompt Writer accordingly rewrites the global prompt $P_0$ and the affected  $\{P_i\}$, and the backend selectively regenerates only the impacted pages. The text agent follows the same principle, updating narration only when a corresponding $S_i$ has changed. This design guarantees that edits propagate precisely and only where intended.



\noindent\textbf{Consistency Critic agent.}
After regeneration, a Consistency Critic agent verifies whether the updated content aligns with the current story state $S$ and, when necessary, neighbouring pages. Using multimodal reasoning over text and images, the Critic checks for attribute mismatches against $C$, missing or mislocated elements relative to $W$ and $S_i$, and layout violations. It then emits structured feedback suggesting minimal corrections to $S_j$ when discrepancies are detected. If accepted, the system updates $S_j$ and re-executes the generation pipeline for that page. This feedback loop allows \textit{StoryState} to detect and repair cross-page inconsistencies without model retraining. Since all control is mediated through structured prompts, the framework remains compatible with diverse T2I backends and LLMs.

\begin{figure*}[t]
  \centering
  \includegraphics[width=1.0\linewidth]{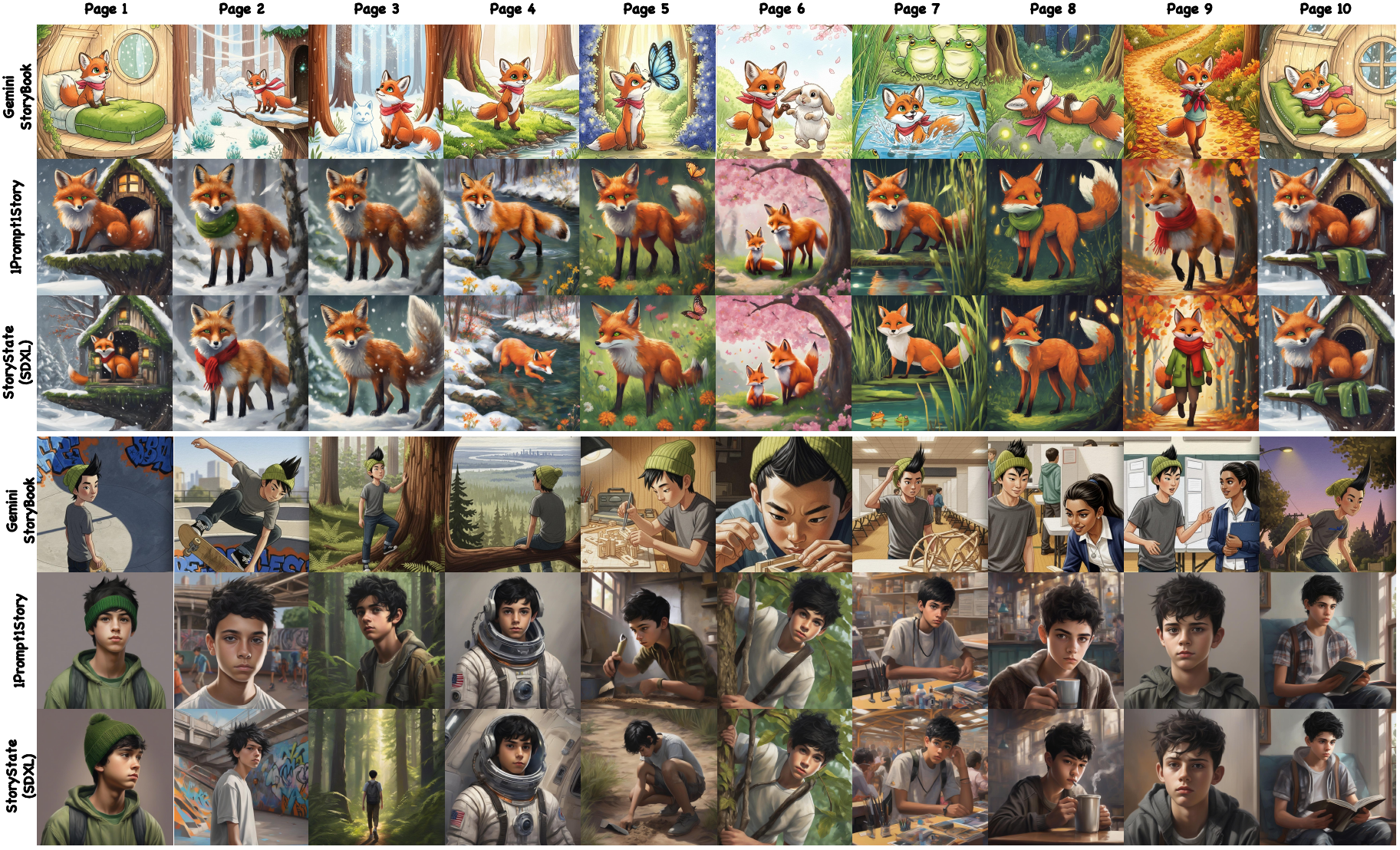}
  \caption{Qualitative comparison across story editing frameworks. Each row shows 10 pages generated by Gemini Storybook, 1Prompt1Story, or \textit{StoryState}. Gemini requires full regeneration on edits; 1Prompt1Story improves consistency but limits diversity in actions and poses; \textit{StoryState} enables localized edits while preserving identity and visual variation across pages.}
  \label{fig:qual}
\end{figure*}

\section{Experiments}

\subsection{Implementation Details}


\noindent\textbf{Dataset.}
We construct a dataset of 192 ten-page illustrated storybooks following the ConsiStory+ benchmark protocol~\cite{liu2025onepromptonestory}. Each story is generated using GPT-5.2, based on a prompt that specifies a main character, a target visual style, and a concise narrative outline. For each base storybooks, we further define multiple page-level edit requests to mimic realistic revision scenarios. This allows the evaluation of iterative editing performance. Representative stories and visual examples are provided in the Appendix \ref{sec:storybook_dataset}.

\noindent\textbf{Settings.}
We compare three methods:
(i) \textit{Gemini Storybook} \cite{google2025storybook}, the official one-click generation tool using Gemini, which does not support iterative edits~\cite{google2025storybookhelp};
(ii) \textit{1Prompt1Story} \cite{liu2025onepromptonestory}, a training-free T2I pipeline adapted to our prompts format; and
(iii) \textit{StoryState}.
All methods receive the same base prompt and target page count. Gemini Storybook is evaluated only for consistency, while 1Prompt1Story and \textit{StoryState} are further evaluated on interactive editing. For \textit{StoryState}, we initialize the state $S$, generate an story, and perform edits by modifying relevant components in $S$ and regenerating the affected pages.

\noindent\textbf{Metrics.}
We report:
(i) \textit{Visual Consistency}, measured as the average cosine similarity of CLIP embeddings between adjacent pages (higher is better);
(ii) \textit{Pages Changed}, the average number of image pages modified after a single edit (lower is better); and
(iii) \textit{User Effort}, captured via user interaction turns and wall-clock time per edit (lower is better).
In the user study we additionally collect 5-point Likert ratings on perceived consistency and control.

\subsection{Comparison with Baselines}

\noindent \textbf{Quantitative comparison.}
As shown in Table~\ref{tab:main}, Gemini Storybook serves as a one-click reference and achieves the highest consistency score (0.89), but it does not support iterative editing, and is therefore no t evaluated on interactivity metrics.
Among the two interactive methods, \textit{StoryState} outperforms 1Prompt1Story in consistency (0.83 vs.\ 0.78) while requiring fewer pages changed (1.6 vs.\ 4.5), fewer user turns (3.1 vs.\ 4.3), and  less time per edit (74\,s vs.\ 96\,s). These results show the benefits of explicit state representation and agent-based prompt control for localized and efficient editing.

\noindent \textbf{Quantitative comparison.}
Figure~\ref{fig:qual} compares stories generated by the three models two examples (Prompts are provided in Appendix~\ref{sec:comparison_prompt}). Gemini Storybook applies edits by regenerating the entire book, often causing unintended changes in unrelated pages. 
1Prompt1Story improves identity consistency but exhibits limited diversity in actions and poses. 
In contrast, \textit{StoryState} preserves a consistent character identity while providing varied poses and scene compositions across pages. its state-based control allows targeted edits to specific page states without unintentionally modifying the rest pages.


\subsection{User Study}

To evaluate perceived visual quality and editing controllability, we conduct a user study with $N=100$ participants (non-experts familiar with illustrated books). For each story prompt, we generate three versions using Gemini Storybook, 1Prompt1Story, and \textit{StoryState}. Participants are shown all three versions and asked:
(i) which story appears more consistent across pages, and
(ii) which system offers greater control during editing.
Participants also rate the overall quality of each story on a 5-point Likert scale. In this section, we focus on pairwise preferences for consistency and control.

\begin{figure}[t]
    \centering
    \includegraphics[width=\linewidth]{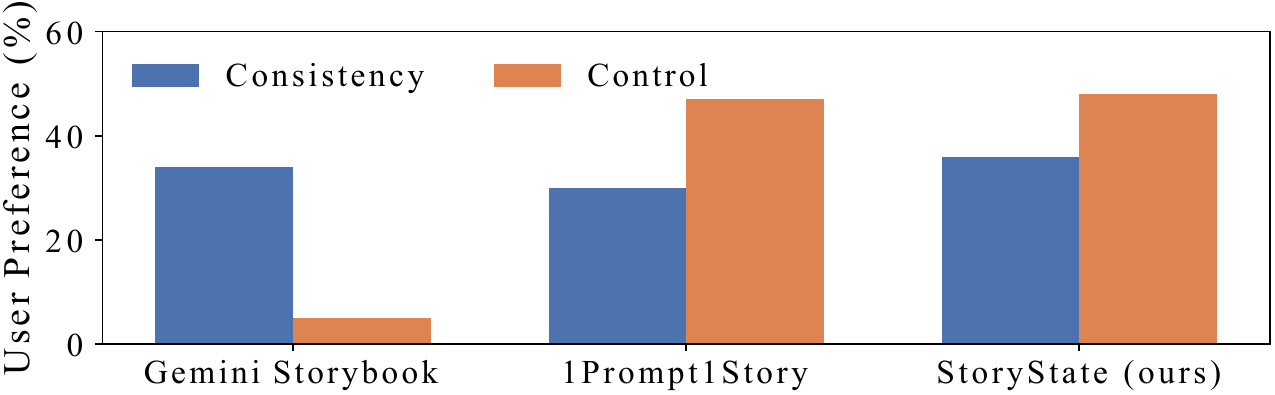}
    \caption{User study comparing consistency and controllability across three story editing systems. \textit{StoryState} is preferred for controllability (48\%) and achieves the highest consistency preference (36\%). See Table~\ref{tab:user} for detailed results.}
    \label{fig:userstudy}
\end{figure}

As shown in Figure~\ref{fig:userstudy}, \textit{StoryState} is preferred in $36\%$ of the comparisons for consistency, slightly higher than Gemini Storybook ($34\%$) and 1Prompt1Story ($30\%$) (see Table~\ref{tab:user}). For perceived control, \textit{StoryState} again leads with $48\%$, closely followed by 1Prompt1Story ($47\%$), while Gemini Storybook is selected in only $5\%$ of the cases. These results indicate that participants find \textit{StoryState} the most controllable system without compromising consistency. In particular, the strong performance of both \textit{StoryState} and 1Prompt1Story on control highlights the advantages of agent-driven editing and structured prompts over Gemini's one-shot interface.

\begin{table}[t]
  \centering
  \caption{System-level comparison across consistency and editing efficiency. \textit{StoryState} achieves high editing efficiency with fewer pages changed, turns, and time per edit, while maintaining high consistency. Gemini Storybook scores highest on consistency but lacks interactive editing capability.}
  \label{tab:main}
  \resizebox{1.0\linewidth}{!}{
  \begin{tabular}{lcccc}
    \toprule
    \textbf{Method} & \textbf{Consistency}$\uparrow$ & \textbf{Pages changed}$\downarrow$ & \textbf{Turns}$\downarrow$ & \textbf{Time (s)}$\downarrow$ \\
    \midrule
    Gemini Storybook   & \textbf{0.89} & - & - & - \\
    1Prompt1Story      & 0.78 & 4.5 & 4.3 & 96 \\
    \textit{StoryState} (ours)  & 0.83 & \textbf{1.6} & \textbf{3.1} & \textbf{74} \\
    \bottomrule
  \end{tabular}
  }
\end{table}


\begin{table}[t]
  \centering
  \caption{Ablation study of \textit{StoryState} components. Explicit state representation and page-level regeneration significantly reduce pages changed and user effort. Removing the Consistency Critic has minor impact but degrades consistency.}
  \label{tab:ablation}
  \resizebox{1.0\linewidth}{!}{
  \begin{tabular}{lcccc}
    \toprule
    \textbf{Variant} & \textbf{Consistency}$\uparrow$ & \textbf{Pages changed}$\downarrow$ & \textbf{Turns}$\downarrow$ & \textbf{Time (s)}$\downarrow$ \\
    \midrule
    Full \textit{StoryState} & 0.83 & \textbf{1.6} & \textbf{3.1} & \textbf{74} \\
    w/o explicit state       & 0.78 & 4.4 & 4.2 & 95 \\
    w/o page-level regen.    & 0.83 & 9.8 & 3.0 & 132 \\
    w/o Consistency Critic   & 0.81 & 1.7 & 3.1 & 75 \\
    \bottomrule
  \end{tabular}
  }
\end{table}




\begin{table}[t]
  \centering
  \caption{Comparison of story editing framework structures. \textit{StoryState} introduces an explicit, persistent state and supports per-page editing, unlike version-based/prompt-only systems.}
  \resizebox{0.95\linewidth}{!}{
  \begin{tabular}{lcc}
    \toprule
    \textbf{Framework} & \textbf{Explicit Story State} & \textbf{Edit Unit} \\
    \midrule
    Gemini Storybook & \xmark & Version \\
    ConsiStory / StoryDiffusion & \xmark & - \\
    1Prompt1Story & \xmark & Prompt \\
    \textit{StoryState} (ours) & \cmark & Page \\
    \bottomrule
  \end{tabular}
  }
  \label{tab:discussion}
\end{table}

\subsection{Ablation Study}

We conduct an ablation study to isolate the impact of three key components in \textit{StoryState}: (i) removing the explicit story-state representation and applying edits via prompt-only regeneration; (ii) disabling page-level regeneration while retaining the state; and (iii) omitting the Consistency Critic.

As Table~\ref{tab:ablation} shows, removing the explicit state leads to substantially more pages being changed and increased user effort, indicating the importance of a persistent, structured representation for localized editing. Disabling page-level regeneration preserves consistency but leads to near-complete story regeneration on each edit, sharply increasing both pages changed and editing time. Excluding the Consistency Critic yields similar quantitative scores but produces more frequent identity and layout violations in qualitative inspection, especially over successive edits. 

These findings confirm that the gains of \textit{StoryState} arise from the combination of persistent state tracking, targeted page-level updates, and structured consistency checking, not merely prompt engineering.


\section{Discussion}

\textit{StoryState} rethinks how storybooks are generated and edited by shifting from prompt-only or version-based workflows to a structured, state-driven paradigm.
Unlike existing systems that operate on opaque prompts or regenerate entire stories for minor edits, \textit{StoryState} externalizes the narrative backbone (characters, global settings, and per-page details) into a persistent and editable representation (Table~\ref{tab:discussion}).

This design makes several core operations such as identity preservation, fine-grained editing, and consistency checking not only feasible but also transparent and controllable. In doing so, \textit{StoryState} reframes story authoring from one-shot generation to incremental, state-centric composition. It offers a middle ground between fully handcrafted pipelines and black-box end-to-end generation, balancing automation and user control in long-form multimodal content creation.



\section{Conclusion}

This paper introduced \textit{StoryState}, an agent-based state controller for consistent and editable storybooks. By explicitly modeling characters, settings, and page-level scenes as structured, editable state, and coupling this with prompt-driven generation, \textit{StoryState} supports precise page-level edits while maintaining cross-page visual consistency. Experiments indicate that \textit{StoryState} reduces unintended changes and user effort while remaining model-agnostic. Beyond storybooks, the state-based control paradigm offers a general approach for managing long-horizon multimodal generation with localized edits. Future work includes extending the state to richer narrative structure, finer-grained object- or region-level edits, and applying StoryState-style orchestration strategy to video generation and interactive media.

\clearpage
\bibliographystyle{IEEEbib}
\bibliography{main}

@misc{google2025storybook,
  title        = {Gemini Storybook --- for the stories only you could imagine},
  author       = {{Google}},
  year         = {2025},
  howpublished = {\url{https://gemini.google/overview/storybook/}},
  note         = {Accessed: 2025-11-15}
}

@misc{google2025storybookblog,
  title        = {Create personal illustrated storybooks in the Gemini app},
  author       = {{Google}},
  year         = {2025},
  howpublished = {\url{https://blog.google/products/gemini/storybooks/}},
  note         = {Accessed: 2025-11-15}
}

@misc{google2025storybookhelp,
  title        = {Create an illustrated storybook in Gemini Apps},
  author       = {{Google}},
  year         = {2025},
  howpublished = {\url{https://support.google.com/gemini/answer/16434396}},
  note         = {Help Center article, accessed: 2025-11-15}
}

@article{theverge2025storybook,
  author  = {Peters, Jay},
  title   = {Google Gemini can now create AI-generated bedtime stories},
  journal = {The Verge},
  year    = {2025},
  note    = {Online article, accessed: 2025-11-15}
}

@article{tewel2024consistory,
  title   = {Training-Free Consistent Text-to-Image Generation},
  author  = {Tewel, Yoad and Kaduri, Omri and Gal, Rinon and Kasten, Yoni and Wolf, Lior and Chechik, Gal and Atzmon, Yuval},
  journal = {ACM Transactions on Graphics},
  year    = {2024}
}

@article{zhou2024storydiffusion,
  title   = {StoryDiffusion: Consistent Self-Attention for Long-Range Image and Video Generation},
  author  = {Zhou, Yupeng and Zhou, Daquan and Cheng, Ming{-}Ming and Feng, Jiashi and Hou, Qibin},
  journal = {NeurIPS},
  year    = {2024}
}

@inproceedings{liu2025onepromptonestory,
  author       = {Tao Liu and Kai Wang and Senmao Li and Joost {van de Weijer}
                  and Fahad Shahbaz Khan and Shiqi Yang and Yaxing Wang
                  and Jian Yang and Ming{-}Ming Cheng},
  title        = {One-Prompt-One-Story: Free-Lunch Consistent Text-to-Image
                  Generation Using a Single Prompt},
  booktitle    = {ICLR},
  year         = {2025},
}

@inproceedings{gal2022textual,
  title     = {An Image is Worth One Word: Personalizing Text-to-Image Generation using Textual Inversion},
  author    = {Gal, Rinon and Alaluf, Yuval and Atzmon, Yuval and Patashnik, Or and Bermano, Amit H. and Chechik, Gal and Cohen-Or, Daniel},
  booktitle = {ICLR},
  year      = {2023}
}

@inproceedings{ruiz2023dreambooth,
  title     = {DreamBooth: Fine Tuning Text-to-Image Diffusion Models for Subject-Driven Generation},
  author    = {Ruiz, Nataniel and Li, Yuanzhen and Jampani, Varun and Pritch, Yael and Rubinstein, Michael and Aberman, Kfir},
  booktitle = {CVPR},
  year      = {2023},
  pages     = {22500--22510}
}

@inproceedings{chen2023suti,
  title     = {Subject-driven Text-to-Image Generation via Apprenticeship Learning},
  author    = {Chen, Wenhu and Hu, Hexiang and Li, Yandong and Ruiz, Nataniel and Jia, Xuhui and Chang, Ming-Wei and Cohen, William W.},
  booktitle = {NeurIPS},
  year      = {2023}
}

@inproceedings{avrahami2024chosen,
  title     = {The Chosen One: Consistent Characters in Text-to-Image Diffusion Models},
  author    = {Avrahami, Omri and Hertz, Amir and Vinker, Yael and Arar, Moab and Fruchter, Shlomi and Fried, Ohad and Cohen-Or, Daniel and Lischinski, Dani},
  booktitle = {ACM SIGGRAPH},
  year      = {2024}
}

@article{zhou2024storymaker,
  title   = {StoryMaker: Towards Holistic Consistent Characters in Text-to-Image Generation},
  author  = {Zhou, Zhengguang and Li, Jing and Li, Huaxia and Chen, Nemo and Tang, Xu},
  journal = {arXiv preprint arXiv:2409.12576},
  year    = {2024}
}

@inproceedings{gong2023interactive,
  title = {Interactive Video Generation with Latent Diffusion Models},
  author = {Gong, Jiaqi and Zhang, Shiwei and Yuan, Hangjie and  others},
  booktitle = {Proceedings of the IEEE/CVF Conference on Computer Vision and Pattern Recognition (CVPR)},
  year = {2023}
}

@article{lin2026narratology,
  title = {Narratology meets text-to-image: a survey of consistency in {AI} generated storybook illustrations},
  author = {Lin, Zhedong and Wang, Zhongsheng and Liu, Qian and Zhang, Xinyu and Liu, Jiamou},
  journal = {Artificial Intelligence Review},
  year = {2026},
  publisher = {Springer},
  doi = {10.1007/s10462-025-11482-6}
}

@inproceedings{lu2025multimodalplan,
  title = {Enhance Multimodal Consistency and Coherence for Text-Image Plan Generation},
  author = {Lu, Xiaoxin and Zhang, Haoran and Zhang, Yusen and Zhang, Rui},
  booktitle = {Findings of the Association for Computational Linguistics (ACL)},
  year = {2025}
}

@article{dong2025vista,
  title = {{ViSTA}: Visual Storytelling using Multi-modal Adapters for Text-to-Image Diffusion Models},
  author = {Dong, Sibo and Shaheen, Ismail and Shen, Maggie and Mallick, Rupayan and Bargal, Sarah Adel},
  journal = {arXiv preprint arXiv:2506.12198},
  year = {2025}
}

@article{he2025dreamstory,
  title={DreamStory: Open-Domain Story Visualization by LLM-Guided Multi-Subject Consistent Diffusion},
  author={He, Huiguo and Wang, Guanying and Zhang, Zhicheng and Zhang, Cheng and Li, Han and Wang, Shiqi and Ling, Nam and Yu, Junkai},
  journal={IEEE Transactions on Pattern Analysis and Machine Intelligence},
  year={2025},
  publisher={IEEE}
}

@article{mao2024storyadapter,
  title={Story-Adapter: A Training-free Iterative Framework for Long Story Visualization},
  author={Mao, Jiawei and He, Huiguo and Wang, Guanying and Yu, Junkai and Ling, Nam},
  journal={arXiv preprint arXiv:2410.06244},
  year={2024}
}

@article{kwon2025dreamingcomics,
  title={DreamingComics: A Story Visualization Pipeline via Subject and Layout Customized Generation using Video Models},
  author={Kwon, Patrick and Chen, Chen},
  journal={arXiv preprint arXiv:2512.01686},
  year={2025}
}

@article{li2025consistentstory,
  title={Consistent Story Generation with Asymmetry Zigzag Sampling},
  author={Li, Mingxiao and Ning, Mang and Moens, Marie-Francine},
  journal={arXiv preprint arXiv:2506.09612},
  year={2025}
}

@article{wang2025storyanchors,
  title={StoryAnchors: Long-form Story Visualization with Consistent Identity and Layout Anchoring},
  author={Wang, Guanying and He, Huiguo and Mao, Jiawei and Zhang, Zhicheng and Ling, Nam and Yu, Junkai},
  journal={arXiv preprint arXiv:2501.03456},
  year={2025}
}

@inproceedings{yu2025cotextor,
  title={CoTextor: Training-Free Modular Multilingual Text Editing via Layered Disentanglement and Depth-Aware Fusion},
  author={Yu, Zhenyu and Idris, Mohd Yamani Idna and Wang, Pei and Qureshi, Rizwan},
  booktitle={The Thirty-ninth Annual Conference on Neural Information Processing Systems Creative AI Track: Humanity},
  year={2025}
}

\clearpage
\appendix
\onecolumn

\renewcommand{\thesection}{A.\arabic{section}}
\renewcommand{\thefigure}{A.\arabic{figure}}
\renewcommand{\thetable}{A.\arabic{table}}

\section{Storybook Dataset}
\label{sec:storybook_dataset}

Listing~\ref{lst:dataset-example} presents a concrete example from the \emph{Storybook Dataset}. Each data instance represents a story composed of ten frames, including a global identity prompt (\texttt{id\_prompt}) and a sequence of frame-level prompts (\texttt{frame\_prompt\_list}). The identity prompt specifies the core visual attributes of the main subject and remains fixed across all frames, while each frame prompt describes a coherent scene with varying actions, environments, or narrative states.

The dataset is designed to support long-horizon visual storytelling by explicitly enforcing identity consistency and structured temporal progression. As illustrated in Listing~\ref{lst:dataset-example}, the frame prompts are organized to follow a natural narrative flow, including scene introduction, environmental development, and resolution. This structured design facilitates controlled multi-frame generation and enables systematic evaluation of identity preservation, temporal coherence, and prompt--image alignment in story-based image generation tasks.

\begin{lstlisting}[
  language=yaml,
  caption={Example of the dataset.},
  label={lst:dataset-example},
  breaklines=true,
  breakatwhitespace=true,
  columns=fullflexible,
  basicstyle=\ttfamily\small,
  showstringspaces=false
]

--id_prompt "A fiery and majestic illustration of A phoenix with bright orange feathers." 
--frame_prompt_list 
  "a phoenix with bright orange feathers, rising from a fiery ashes, introducing the scene, captured as a vivid and coherent moment with cinematic lighting, clear spatial context, and strong visual continuity" 
  "a phoenix with bright orange feathers, soaring through a glowing sky, introducing the scene, captured as a vivid and coherent moment with cinematic lighting, clear spatial context, and strong visual continuity" 
  "a phoenix with bright orange feathers, perching on a mountain peak, developing the environment, captured as a vivid and coherent moment with cinematic lighting, clear spatial context, and strong visual continuity" 
  "a phoenix with bright orange feathers, singing a haunting melody, developing the environment, captured as a vivid and coherent moment with cinematic lighting, clear spatial context, and strong visual continuity" 
  "a phoenix with bright orange feathers, igniting flames with its wings, developing the environment, captured as a vivid and coherent moment with cinematic lighting, clear spatial context, and strong visual continuity" 
  "a phoenix with bright orange feathers, rising from a fiery ashes, developing the environment, captured as a vivid and coherent moment with cinematic lighting, clear spatial context, and strong visual continuity" 
  "a phoenix with bright orange feathers, soaring through a glowing sky, bringing the scene to a calm resolution, captured as a vivid and coherent moment with cinematic lighting, clear spatial context, and strong visual continuity" 
  "a phoenix with bright orange feathers, perching on a mountain peak, bringing the scene to a calm resolution, captured as a vivid and coherent moment with cinematic lighting, clear spatial context, and strong visual continuity" 
  "a phoenix with bright orange feathers, singing a haunting melody, bringing the scene to a calm resolution, captured as a vivid and coherent moment with cinematic lighting, clear spatial context, and strong visual continuity" 
  "a phoenix with bright orange feathers, igniting flames with its wings, bringing the scene to a calm resolution, captured as a vivid and coherent moment with cinematic lighting, clear spatial context, and strong visual continuity" 

--id_prompt "A vibrant and striking portrait of A zebra with black and white stripes." 
--frame_prompt_list 
  "a zebra with black and white stripes, grazing alongside a river, introducing the scene, captured as a vivid and coherent moment with cinematic lighting, clear spatial context, and strong visual continuity" 
  "a zebra with black and white stripes, running in a herd across the plains, introducing the scene, captured as a vivid and coherent moment with cinematic lighting, clear spatial context, and strong visual continuity" 
  "a zebra with black and white stripes, resting under the shade of an acacia tree, developing the environment, captured as a vivid and coherent moment with cinematic lighting, clear spatial context, and strong visual continuity" 
  "a zebra with black and white stripes, crossing a dusty path in the wild, developing the environment, captured as a vivid and coherent moment with cinematic lighting, clear spatial context, and strong visual continuity" 
  "a zebra with black and white stripes, protecting its young in the savannah, developing the environment, captured as a vivid and coherent moment with cinematic lighting, clear spatial context, and strong visual continuity" 
  "a zebra with black and white stripes, grazing alongside a river, developing the environment, captured as a vivid and coherent moment with cinematic lighting, clear spatial context, and strong visual continuity" 
  "a zebra with black and white stripes, running in a herd across the plains, bringing the scene to a calm resolution, captured as a vivid and coherent moment with cinematic lighting, clear spatial context, and strong visual continuity" 
  "a zebra with black and white stripes, resting under the shade of an acacia tree, bringing the scene to a calm resolution, captured as a vivid and coherent moment with cinematic lighting, clear spatial context, and strong visual continuity" 
  "a zebra with black and white stripes, crossing a dusty path in the wild, bringing the scene to a calm resolution, captured as a vivid and coherent moment with cinematic lighting, clear spatial context, and strong visual continuity" 
  "a zebra with black and white stripes, protecting its young in the savannah, bringing the scene to a calm resolution, captured as a vivid and coherent moment with cinematic lighting, clear spatial context, and strong visual continuity" 

--id_prompt "A sleek and fast depiction of A cheetah with sharp eyes." 
--frame_prompt_list 
  "a cheetah with sharp eyes, sprinting across the savannah, introducing the scene, captured as a vivid and coherent moment with cinematic lighting, clear spatial context, and strong visual continuity" 
  "a cheetah with sharp eyes, stalking a gazelle in the grass, introducing the scene, captured as a vivid and coherent moment with cinematic lighting, clear spatial context, and strong visual continuity" 
  "a cheetah with sharp eyes, relaxing in the shade under a tree, developing the environment, captured as a vivid and coherent moment with cinematic lighting, clear spatial context, and strong visual continuity" 
  "a cheetah with sharp eyes, marking territory with its scent, developing the environment, captured as a vivid and coherent moment with cinematic lighting, clear spatial context, and strong visual continuity" 
  "a cheetah with sharp eyes, napping after a successful hunt, developing the environment, captured as a vivid and coherent moment with cinematic lighting, clear spatial context, and strong visual continuity" 
  "a cheetah with sharp eyes, sprinting across the savannah, developing the environment, captured as a vivid and coherent moment with cinematic lighting, clear spatial context, and strong visual continuity" 
  "a cheetah with sharp eyes, stalking a gazelle in the grass, bringing the scene to a calm resolution, captured as a vivid and coherent moment with cinematic lighting, clear spatial context, and strong visual continuity" 
  "a cheetah with sharp eyes, relaxing in the shade under a tree, bringing the scene to a calm resolution, captured as a vivid and coherent moment with cinematic lighting, clear spatial context, and strong visual continuity" 
  "a cheetah with sharp eyes, marking territory with its scent, bringing the scene to a calm resolution, captured as a vivid and coherent moment with cinematic lighting, clear spatial context, and strong visual continuity" 
  "a cheetah with sharp eyes, napping after a successful hunt, bringing the scene to a calm resolution, captured as a vivid and coherent moment with cinematic lighting, clear spatial context, and strong visual continuity" 

--id_prompt "A massive and majestic depiction of A walrus with large tusks." 
--frame_prompt_list 
  "a walrus with large tusks, lounging on an ice floe, introducing the scene, captured as a vivid and coherent moment with cinematic lighting, clear spatial context, and strong visual continuity" 
  "a walrus with large tusks, diving for clams in the sea, introducing the scene, captured as a vivid and coherent moment with cinematic lighting, clear spatial context, and strong visual continuity" 
  "a walrus with large tusks, bellowing from the shore, developing the environment, captured as a vivid and coherent moment with cinematic lighting, clear spatial context, and strong visual continuity" 
  "a walrus with large tusks, resting near a snowy coastline, developing the environment, captured as a vivid and coherent moment with cinematic lighting, clear spatial context, and strong visual continuity" 
  "a walrus with large tusks, swimming gracefully in the ocean, developing the environment, captured as a vivid and coherent moment with cinematic lighting, clear spatial context, and strong visual continuity" 
  "a walrus with large tusks, lounging on an ice floe, developing the environment, captured as a vivid and coherent moment with cinematic lighting, clear spatial context, and strong visual continuity" 
  "a walrus with large tusks, diving for clams in the sea, bringing the scene to a calm resolution, captured as a vivid and coherent moment with cinematic lighting, clear spatial context, and strong visual continuity" 
  "a walrus with large tusks, bellowing from the shore, bringing the scene to a calm resolution, captured as a vivid and coherent moment with cinematic lighting, clear spatial context, and strong visual continuity" 
  "a walrus with large tusks, resting near a snowy coastline, bringing the scene to a calm resolution, captured as a vivid and coherent moment with cinematic lighting, clear spatial context, and strong visual continuity" 
  "a walrus with large tusks, swimming gracefully in the ocean, bringing the scene to a calm resolution, captured as a vivid and coherent moment with cinematic lighting, clear spatial context, and strong visual continuity" 

--id_prompt "A bold and striking illustration of A zebra with black and white stripes." 
--frame_prompt_list 
  "a zebra with black and white stripes, grazing under the sun in the savannah, introducing the scene, captured as a vivid and coherent moment with cinematic lighting, clear spatial context, and strong visual continuity" 
  "a zebra with black and white stripes, running with a herd through the grasslands, introducing the scene, captured as a vivid and coherent moment with cinematic lighting, clear spatial context, and strong visual continuity" 
  "a zebra with black and white stripes, standing near a watering hole, developing the environment, captured as a vivid and coherent moment with cinematic lighting, clear spatial context, and strong visual continuity" 
  "a zebra with black and white stripes, crossing a river in the wild, developing the environment, captured as a vivid and coherent moment with cinematic lighting, clear spatial context, and strong visual continuity" 
  "a zebra with black and white stripes, trotting beside a wildebeest, developing the environment, captured as a vivid and coherent moment with cinematic lighting, clear spatial context, and strong visual continuity" 
  "a zebra with black and white stripes, grazing under the sun in the savannah, developing the environment, captured as a vivid and coherent moment with cinematic lighting, clear spatial context, and strong visual continuity" 
  "a zebra with black and white stripes, running with a herd through the grasslands, bringing the scene to a calm resolution, captured as a vivid and coherent moment with cinematic lighting, clear spatial context, and strong visual continuity" 
  "a zebra with black and white stripes, standing near a watering hole, bringing the scene to a calm resolution, captured as a vivid and coherent moment with cinematic lighting, clear spatial context, and strong visual continuity" 
  "a zebra with black and white stripes, crossing a river in the wild, bringing the scene to a calm resolution, captured as a vivid and coherent moment with cinematic lighting, clear spatial context, and strong visual continuity" 
  "a zebra with black and white stripes, trotting beside a wildebeest, bringing the scene to a calm resolution, captured as a vivid and coherent moment with cinematic lighting, clear spatial context, and strong visual continuity" 
\end{lstlisting}

\section{Prompt of Comparison Story}
\label{sec:comparison_prompt}

This section provides the prompts used for the comparison methods in our experiments (see Listing \ref{lst:comparison-prompt}). To ensure a fair and reproducible evaluation, all baseline methods are conditioned on the same story prompts, which define the main character identity and the sequence of narrative scenes. These prompts are directly adopted as inputs for the corresponding comparison pipelines, without any manual tuning or method-specific modification.

The comparison prompts follow the same structured format as those used in our method, consisting of a fixed identity prompt and a sequence of frame-level prompts. By standardizing the prompt content across all methods, we isolate the performance differences to the modeling and generation mechanisms rather than prompt engineering. The complete prompt examples are included in this section to facilitate transparency, reproducibility, and future benchmarking.

\begin{lstlisting}[
  language=yaml,
  caption={Example of the dataset.},
  label={lst:comparison-prompt},
  breaklines=true,
  breakatwhitespace=true,
  columns=fullflexible,
  basicstyle=\ttfamily\small,
  showstringspaces=false
]

--id_prompt "Fira, a curious fox with orange fur, bright green eyes, and a small red scarf around her neck." 
--frame_prompt_list 
  "waking up in her cozy treehouse as the first snowflakes of winter fall outside the window" 
  "stepping into a snowy clearing, her red scarf fluttering as she explores the quiet forest" 
  "building a small snow fox and watching magical snow sprites dance in the air" 
  "walking along a stream as the snow begins to melt and tiny spring flowers start to bloom" 
  "meeting a glowing butterfly that leads her to a hidden meadow filled with blooming flowers" 
  "playing with other animals beneath cherry blossom trees in a gentle spring breeze" 
  "jumping into a cool pond surrounded by glowing reeds and friendly frogs in summer" 
  "watching fireflies swirl above her during a warm summer night in the magical forest" 
  "walking along a golden path as autumn leaves fall, her red scarf blending into the trees" 
  "curling up in her treehouse as winter returns, gazing at the falling snow with a peaceful smile"

--id_prompt "A hyper-realistic digital painting of a teenage boy with short, spiky black hair, a slight build, and dark brown eyes." 
--frame_prompt_list 
  "wearing a green knitted hat, standing quietly as soft light falls across his face" 
  "spending time at an urban skatepark surrounded by concrete ramps and graffiti walls" 
  "walking through a dense forest with tall trees and filtered sunlight" 
  "dressed as an astronaut, appearing ready for an imaginative journey beyond Earth" 
  "kneeling on the ground and carefully digging with a small trowel" 
  "climbing a tree and gripping the branches as leaves rustle around him" 
  "attending a lively science fair filled with posters, models, and curious visitors" 
  "holding a warm drink as steam rises gently in a quiet moment" 
  "painting a picture on a canvas, fully absorbed in the creative process" 
  "sitting calmly while reading a book, lost in thought and imagination"

  
\end{lstlisting}

\section{Examples of \textit{StoryState}}
\label{sec:examples_storystate}

\begin{figure}[h]
    \centering
    \includegraphics[width=1\linewidth]{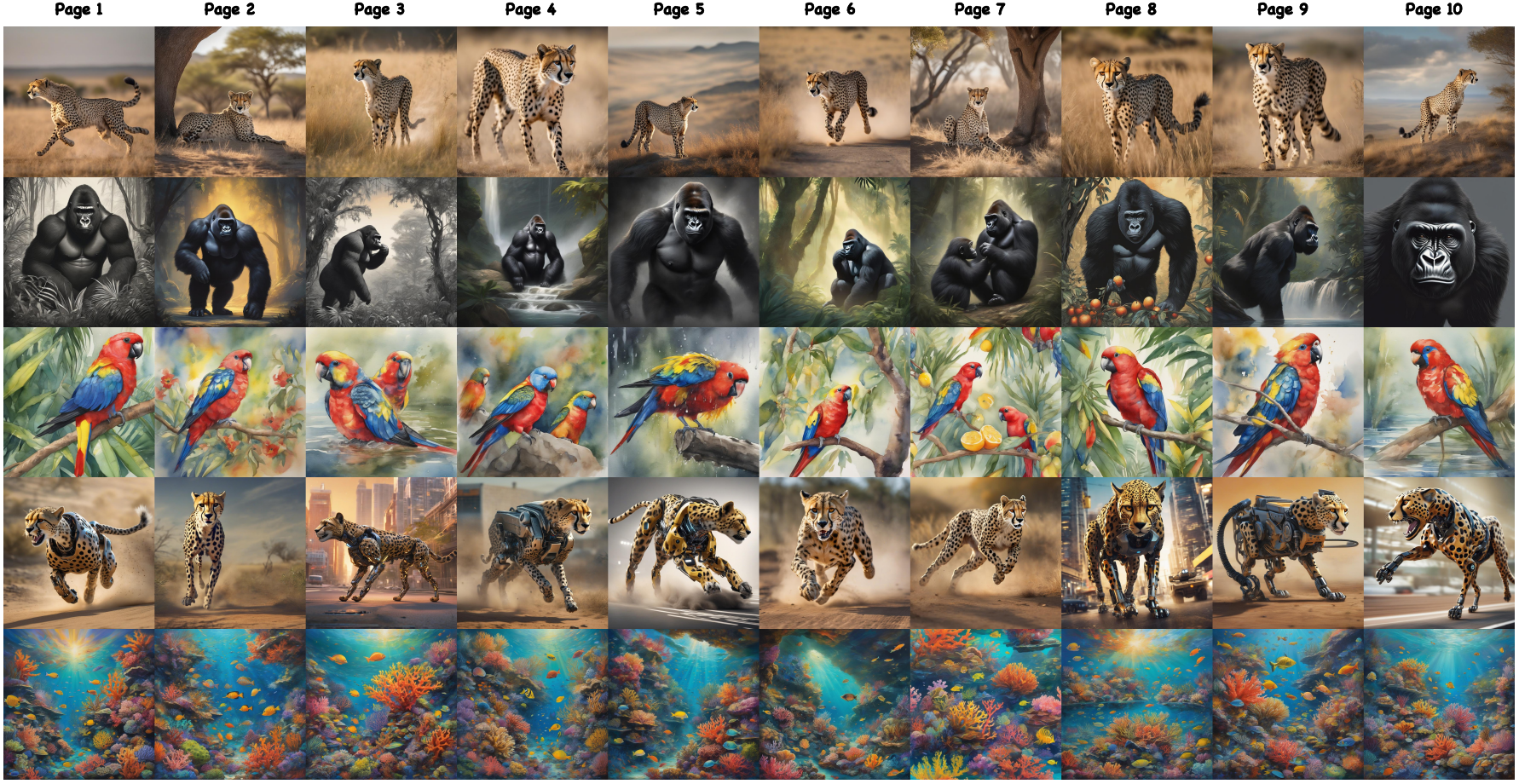}
    \caption{Examples of \textit{StoryState} generated results.}
    \label{fig_examples_storystate1}
\end{figure}

\begin{figure}[h]
    \centering
    \includegraphics[width=1\linewidth]{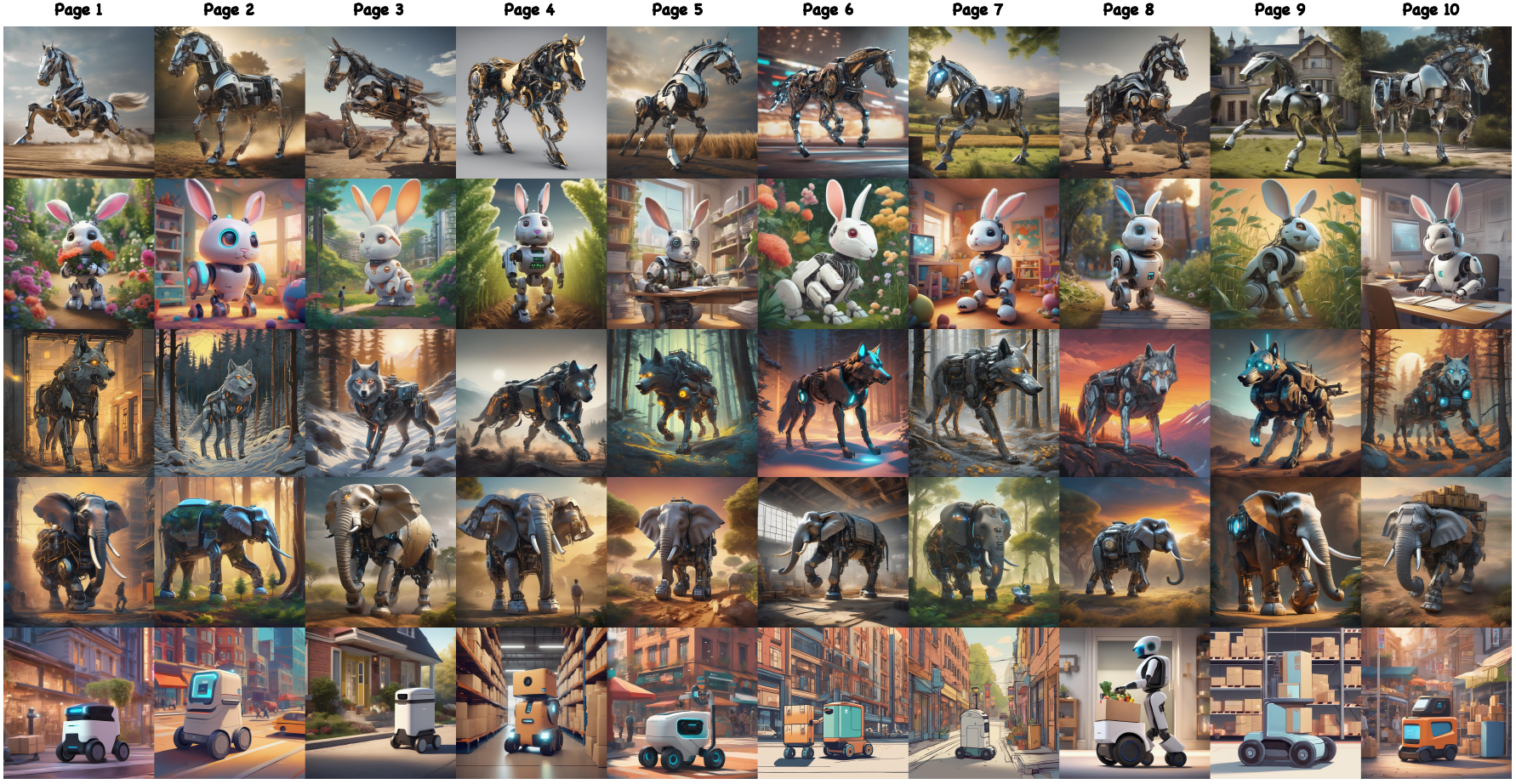}
    \caption{Examples of \textit{StoryState} generated results.}
    \label{fig_examples_storystate2}
\end{figure}

This section presents qualitative examples generated by \textit{StoryState} to illustrate its behavior in long-horizon visual storytelling (see Figure \ref{fig_examples_storystate1} and \ref{fig_examples_storystate2}.).
Each example demonstrates how \textit{StoryState} maintains a consistent representation of the main character while progressively updating the underlying state to reflect changes in actions, environments, and narrative context across frames.
Unlike baseline methods that treat multi-frame generation as independent or weakly coupled processes, \textit{StoryState} explicitly models generation as a state-dependent process.
As shown in the examples, state transitions enable coherent temporal evolution without identity drift or visual degeneration, even in extended story sequences.
These qualitative results complement the quantitative analysis in the main paper and provide intuitive insights into how state-aware inference contributes to stable and controllable story generation.


\clearpage
\section{Additional Tables}
Table~\ref{tab:user} reports the detailed results of the user study on perceived visual consistency and editing controllability. Participants were asked to compare storybooks generated by different systems and select the one that they perceived as more consistent across pages and easier to control during editing. The percentages indicate the proportion of times each system was preferred in pairwise comparisons. As shown in the table, \textit{StoryState} achieves the highest preference rates for both consistency and control, suggesting that explicit story-state representation and agent-based orchestration improve users’ perceived editing experience.

\begin{table}[h]
  \centering
  \caption{User study results. \textbf{Bold} values indicate the best performance.}
  \label{tab:user}
  \begin{tabular}{lcc}
    \toprule
    \textbf{System} & \textbf{Consistency}$\uparrow$ & \textbf{Control}$\uparrow$ \\
    \midrule
    Gemini Storybook      & 34\% & 5\% \\
    1Prompt1Story         & 30\% & 47\% \\
    \textit{StoryState} (ours)     & \textbf{36\%} & \textbf{48\%} \\
    \bottomrule
  \end{tabular}
\end{table}

\end{document}